# A Conditional Mutual Information Analysis of Response Variety in Language Models

R. I. Cukier

*Department of Chemistry, Michigan State University, East Lansing, Michigan 48824-1322, USA*

cukier@chemistry.msu.edu

Abstract

Large language models provide a variety of responses at inference/generation that conventionally depend on a "temperature" parameter. Using a conditional mutual information approach, a quantitative, entropy-based analysis of response variety is given. Three limiting behaviors are defined: arbiter, independent (conditionally) and neutral/deterministic that span no to maximal mutual information. We give examples of each behavior and suggest their application to: "temperature" analysis, language analysis, prompt compression, and knowledge graphs.

## 1. Introduction

Large language models (LLMs), with their autoregressive character, dominate text generation. Based on a "temperature" parameter, LLMs in inference/generative mode may provide a variety of responses to a given prompt. This work provides an analysis method of the variety of responses to a given prompt. It is more expressive than simply setting a temperature. The basis is a conditional version of mutual information [Cover]:

$$I\left(\mathbf{K}^1,\mathbf{K}^2 \mid Q\right) = E_{p(q)} I_q\left(\mathbf{K}^1,\mathbf{K}^2 \mid Q=q\right) = E_{p(q)} E_{p(\mathbf{k}^1,\mathbf{k}^2|q)} \log\left[\frac{p\left(\mathbf{K}^1,\mathbf{K}^2 \mid Q\right)}{p\left(\mathbf{K}^1 \mid Q\right) p\left(\mathbf{K}^2 \mid Q\right)}\right] \quad (1.1.a)$$

with the operational expression

$$I_q\left(\mathbf{K}^1,\mathbf{K}^2 \mid Q=q\right) := \sum_{\mathbf{k}^1}\sum_{\mathbf{k}^2} p\left(\mathbf{k}^1,\mathbf{k}^2 \mid q\right) \log\left[\frac{p\left(\mathbf{k}^1,\mathbf{k}^2 \mid q\right)}{p\left(\mathbf{k}^1 \mid q\right) p\left(\mathbf{k}^2 \mid q\right)}\right] \quad (1.1.b)$$

In Eq. (1.1), $\mathbf{K}^1$ and $\mathbf{K}^2$ are phrases that pre- and post-pend $Q$; the latter thought of as a query, $q \in Q$.

We shall show that

$$0 \le I\left(\mathbf{K}^1, \mathbf{K}^2 \mid Q\right) \le \min\left[H\left(\mathbf{K}^1 \mid Q\right), H\left(\mathbf{K}^2 \mid Q\right)\right] \le H\left(\mathbf{X}\right) \quad \left(\mathbf{K}^1, \mathbf{K}^2, Q \in \mathbf{X}\right) \tag{1.2}$$

with $H\left(\mathbf{X}\right)$ the entropy of the vocabulary.

Without the conditionality, Eq. (1.1) simplifies to the standard mutual information (MI) between the probability of two correlated random variables (RVs) and their product, independent probabilities. Cover and Thomas' book [Cover] also provides the conditional mutual information (CMI) version in Eq. (1.1) and various rewritings of it are used to obtain Eq. (1.2) and to analyze its limiting behaviors. When the CMI is maximum, one learns the most possible about $\mathbf{K}^2$ from knowledge of $\mathbf{K}^1$ (and vice versa), conditional on $Q = q$. When zero, no added information about $\mathbf{K}^2$ due to knowledge of $\mathbf{K}^1$ (and vice versa), conditional on query $Q = q$ is obtained. Based on these limits, we define three instructive classes termed: arbiter, independent (conditionally) and neutral/deterministic. Examples of each class and their application to the four topics: 1) "temperature" analysis, 2) language analysis, 3) prompt compression and 4) knowledge graphs are presented.

## 2. Derivations

### 2.1 General Expressions

The following relies on Cover and Thomas' book [Cover]. For notational simplicity, use "1", "2" and "3" for the RV names. The conditional MI in Eq. (1.1.a), by expansion of numerator and denominator, is

$$I\left(13 \mid 2\right) = H\left(1 \mid 2\right) - H\left(1 \mid 23\right) \tag{2.1}$$

in terms of the corresponding entropies.

Use the entropy decomposition expression

$$H\left(123\right) = H\left(1 \mid 23\right) + H\left(2 \mid 3\right) + H\left(3\right), \tag{2.2}$$

written for $-H\left(1 \mid 23\right)$, in Eq. (2.1) to get

$$I(13|2)=H(1|2)+H(2|3)+H(3)-H(123). \tag{2.3}$$

Now use another version of $H(123)$:

$$H(123)=H(12|3)+H(3). \tag{2.4}$$

Substitute Eq. (2.4) in Eq. (2.3) to eliminate $H(3)$:

$$\begin{aligned} I(13|2)&=H(1|2)+H(2|3)+H(3)-\left[H(12|3)+H(3)\right] \\ \therefore I(13|2)&=H(1|2)+H(2|3)-H(12|3) \end{aligned} \tag{2.5}$$

From Eq. (2.1),

$$I(13|2)\le H(1|2). \tag{2.6}$$

From Eq. (2.5),

$$I(13|2)\le H(1|2)+H(3|2). \tag{2.7}$$

Eq. (2.6) provides a tighter bound than that from Eq. (2.7).

From the 'symmetry' of Eq. (2.6),

$$I(13|2)\le \min\left[H(1|2),H(3|2)\right]\square. \tag{2.8}$$

2.2 Limiting Cases.

Various limiting behaviors are instructive. Throughout, assume that functional relations between given RVs are one-to-one, as is sensible for tokens.

Suppose $1=f(3)$. Then $H(1|23)\underset{1=f(3)}{\rightarrow}0$ because there can be only one value of RV 1 for a specified value of RV 3. Thus, from Eq. (2.1)

$$I(13|2)\underset{1=f(3)}{\rightarrow}H(1|2)\square. \tag{2.9}$$

Eq. (2.9) maximizes Eq. (2.8). Note that for any vocabulary it is straightforward to evaluate the conditional entropy $H(1|2)$.

Suppose $3=f(2)$. Then $H(1|23)\underset{3=f(2)}{\rightarrow}H(1|2)$ because RV 3 provides no added information about RV 2, given a value of RV 2. From Eq. (2.1)

$$I(13|2)=H(1|2)-H(1|23)\underset{3=f(2)}{\rightarrow}H(1|2)-H(1|2)=0\square. \tag{2.10}$$

Lastly, as conditioning can only reduce entropy, Eq. (2.8) has the further limiting behavior that if e.g., the minimum in Eq. (2.8) is $H(1|2)$ then

$$I(1,3|2) \le H(1|2) \le H(1)\square. \tag{2.11}$$

Explicit proofs of the above two limiting cases are provided in Appendix A.

**3. Limit Behavior and Terminology**

With the above limit behaviors, three situations: arbiter, independent (conditionally) and neutral/deterministic can be inferred.

From Eq. (2.9), where $1 = f(3)$, using the terminology and notation of the Introduction,

1) **arbiter**: $I(\mathbf{K}^1, \mathbf{K}^2 | Q) \underset{\mathbf{K}^1 = f(\mathbf{K}^2)}{\rightarrow} H(\mathbf{K}^1 | Q) \le H(\mathbf{K}^1) = H(\mathbf{X})$ (2.12)

where $\{\mathbf{K}^1, \mathbf{K}^2, Q\} \in \mathbf{X}$, with $\mathbf{X}$ the token set of entropy $H(\mathbf{X})$. Here, one can conclude that, for a set of query words, there is a strong chance that for $\mathbf{K}^1$ you "know" what $\mathbf{K}^2$ is. Examples are given in Table 1.

| $\mathbf{K}^1$ | $\mathbf{K}^2$ | $Q = q$ |
|---|---|---|
| the black cat | over the moon | jumped |
| peanut butter | jelly | and |
| bread | butter | and |
| peaches | cream | and |
| I | you | love |
| love | all | conquers |

Table 1. Arbiter examples.

This circumstance should be more likely for longer phrases, such as poetry/lyrics.

Regarding the inequality in Eq. (2.12), note that for many queries there will be $\{\mathbf{K}^1\}$ sets where there is no correlation with a $q$. E.g., $q = [\text{and}]$ with $\mathbf{K}^1$ just about any word, such as

{bread, milk, cat, ….} $\rightarrow [\text{and}]$. For these $q$'s, $I_q\left(\mathbf{K}^1, \mathbf{K}^2 \mid Q=q\right) \rightarrow H\left(\mathbf{K}^1\right)$, the maximum possible entropy, which is the entropy of the vocabulary, as $\mathbf{Q}=\mathbf{K}=\mathbf{V}=\mathbf{X}$.

2) **Independent (conditionally)**: $I\left(\mathbf{K}^1, \mathbf{K}^2 \mid Q=q\right)=0$. (2.13)

Independence results from Eq. (1.1) when $p\left(\mathbf{K}^1, \mathbf{K}^2 \mid Q=q\right)=p\left(\mathbf{K}^1 \mid Q=q\right) p\left(\mathbf{K}^2 \mid Q=q\right)$ where, by definition, $\mathbf{K}^2$ is uninformative about $\mathbf{K}^1$. Examples, with representation $\left(\mathbf{K}^1\right) \rightarrow [Q=q] \rightarrow \left(\mathbf{K}^2\right)$, are:

(he, they, she,…) $\rightarrow$[looked] $\rightarrow$(up, down, sideways,…).

This should be a common occurrence.

3) **Neutral/deterministic:** $I\left(\mathbf{K}^1, \mathbf{K}^2 \mid Q=q\right)=0$. (2.14)

This occurs when $\mathbf{K}^2=f\left(Q=q\right)$, (or $\mathbf{K}^1=f\left(Q=q\right)$), where there is no correlation between prefix and suffix, conditional on a query. We refer to it as neutral as nothing is learned about $\mathbf{K}^2$ when $\mathbf{K}^1$ is determined by a query $Q=q$. We refer to it as deterministic because $p\left(\mathbf{K}^2=\mathbf{k} \mid Q=q\right)=\delta_{\mathbf{k}, \hat{\mathbf{k}}}$ (or $p\left(\mathbf{K}^1=\mathbf{k} \mid Q=q\right)=\delta_{\mathbf{k}, \hat{\mathbf{k}}}$) with $\hat{\mathbf{k}}$ a token. This case should also be a common occurrence. It is easy to distinguish between independent (conditionally**)** and neutral/deterministic where, in both cases $I\left(\mathbf{K}^1, \mathbf{K}^2 \mid Q=q\right)=0$, because for neutral/deterministic $p\left(\mathbf{K}^2=k \mid Q=q\right)=\delta_{\mathbf{k}, \hat{\mathbf{k}}}$ (or $p\left(\mathbf{K}^1=\mathbf{k} \mid Q=q\right)=\delta_{\mathbf{k}, \hat{\mathbf{k}}}$).

Examples of neutral/deterministic with pattern $[Q=q] \rightarrow \left(\mathbf{K}^2\right)$ are shown in Table 2. E.g., for [Carpe] $\rightarrow$( Diem) – a fixed prefix for this query, [Carpe]. That is, there is only one $\mathbf{K}^2$ for this $Q=q$.

| $[Q=q]$ | $\left(\mathbf{K}^2\right)$ |
|---|---|
| Carpe | Diem |
| chit | chat |
| topsy | turvy |
| sink or | swim |
| related | to |

Table 2. neutral/deterministic examples.

Of course, the last example in Table 2 is not precise because, e.g., $\mathbf{K}^2 = (\text{with})$ is also possible. But the others are distinctly deterministic.

**4. Uses**

Four uses of this mutual information are discussed here:

1) **Temperature**. Generically, the above analysis can be viewed as a more specific approach to "temperature$(T)$". In generative mode $T$ is used to span, for low $T$, almost always the same response, to high $T$, with a greater variety of responses. Tasks such as mathematical proofs are appropriate to low $T$ while poetry would be better served by a higher $T$. (Other methods of producing a variety of responses such as "top-K" and "top-P" exist of course. [von Platen])

2) **Language analysis**. E.g., for a "connector" $(Q = q)$ word. If a connector, $q$, is "and" then one expects $I_q\left(\mathbf{K}^1, \mathbf{K}^2 \mid Q = q\right) > 0$ because of phrases such as: (bread) – [and] – (butter), (peaches) – [and] – (cream), (black) – [and] – (white), etc. On the other hand, good candidates for a connector where $I_q\left(\mathbf{K}^1, \mathbf{K}^2 \mid Q = q\right) = 0$ because of independence could be, for examples, (he/they/she/…) – [looked] – (up/down/sideways…).

3) **Prompt compression**. Here, the arbiter $I\left(\mathbf{K}^1, \mathbf{K}^2 \mid Q\right) = H\left(\mathbf{K}^1 \mid Q\right) \le H\left(\mathbf{K}^1\right)$ for $\mathbf{K}^2 = f\left(\mathbf{K}^1\right)$ of Eq. (2.12) is most appropriate. A large CMI implies you learn about $\mathbf{K}^2$ from knowing $\mathbf{K}^1$; so, $\mathbf{K}^1$ can be skipped (and vice versa). For example,

(A hurricane has been announced.) – [Therefore,] – (air traffic will be disrupted.)

This $\mathbf{K}^2$ will, with more than random weight, be preceded by this $\mathbf{K}^1$. Phrases such as:

(black) –[and] –(white)

come up often in texts. The probabilities of such phrases are higher than random implying the arbiter situation.

4) **Knowledge Graphs**. In the context of knowledge graphs (KGs), the association of nodes (entities) with keys, $\mathbf{K}^1$ and $\mathbf{K}^2$, and edges (relationships) as queries, $Q$, is warranted. KGs [Edge] have become important to the systematization of Retrieval Augmented Generation (RAG) [Dalimunthe, Zhao], with ongoing efforts to improve prompt design in Large Language Model inference/generation. Knowledge graphs should also fit the arbiter category because what is desired is $\mathbf{K}^1 = f\left(\mathbf{K}^2\right)$ whereby $I\left(\mathbf{K}^1, \mathbf{K}^2 \mid Q\right) = H\left(\mathbf{K}^1 \mid Q\right) = H\left(\mathbf{K}^2 \mid Q\right)$. In other words, maximum CMI because, if phrases $\mathbf{K}^2$ and $\mathbf{K}^1$ are correlated, there is a lot of mutual information, conditional on $Q = q$. For example, a relationship word such as $q =$[developed] can show a strong connection between two keys as in: (Albert Einstein) → [developed] → (a theory of relativity), as further elaborated in [Dalimunthe]. Relationship words with large CMI (approximately the entropy of a key, conditional on a given query) should help create KGs in an objective manner.

**5. Concluding Remarks**

In this work, we develop a method to analyze the relationship among words and phrases based on the conditional mutual information expression given in Eq. (1.1). Various limits of it follow from the connection between the two keys, $\mathbf{K}^1$ and $\mathbf{K}^2$, or between a key, $\mathbf{K}^1$ or $\mathbf{K}^2$, and the query, $Q = q$. Expressing the conditional mutual information in terms of its constituent entropies provides various limiting expressions. From Eq. 2.9, it is straightforward, given a token set $\mathbf{X}$, to evaluate e.g., $H\left(\mathbf{K}^1 \mid Q = q\right) \to H\left(\mathbf{X} \mid Q = x\right)$ because all tokens are from same alphabet. The non-negativity of this entropy provides strict bounds on the possible CMI values.

Three cases are defined: arbiter, independent (conditionally) and neutral/deterministic that achieve limiting behaviors of the conditional mutual information. They are applied to four tasks: "Temperature" analysis, language analysis, prompt compression, and knowledge graphs. The most interesting cases correspond to the arbiter situation where the CMI is close to the conditional entropy $H\left(\mathbf{X} \mid Q = x\right)$ as it relies on a strong relationship between the keys. As shown, this should be quite useful for prompt compression and knowledge graph/Retrieval Augmented Generation (KG/RAG) purposes.

Finally, with reference to Eq. (1.1), a similar analysis can be conducted for $I\left(Q, \mathbf{K}^1 \mid \mathbf{K}^2\right)$ or $I\left(Q, \mathbf{K}^2 \mid \mathbf{K}^1\right)$. The only distinction is which RV is fixed.

**Appendix A**

Eqs. (2.9) and (2.10) follow from the definitions of mutual information and entropy.

For Eq. (2.9), where $1 = f(3)$:

$$p(x_1, x_3 | x_2) = 1_{x_1 = f(x_3)} p(x_3 | x_2)$$

$$I(x_1, x_3 | x_2) = E\left[\log \frac{p(x_1, x_3 | x_2)}{p(x_1 | x_2) p(x_3 | x_2)}\right] = E\left[\log \frac{1_{x_1 = f(x_3)} p(x_3 | x_2)}{p(x_1 | x_2) p(x_3 | x_2)}\right] \tag{A.1}$$

$$= E\left[\log \frac{1_{x_1 = f(x_3)}}{p(x_1 | x_2)}\right] = E\left[\log \frac{1}{p(f(x_3) | x_2)}\right] = E\left[\log \frac{1}{p(x_1 | x_2)}\right] := H(1|2) \square$$

For Eq. (2.10), where $3 = f(2)$:

$$H(1|23) \underset{3=f(2)}{\rightarrow} -\sum_{x_1} \sum_{x_2} \sum_{x_3} p(x_1 | x_2, x_3) 1_{x_3 = f(x_2)} p(x_2) \log p(x_1, x_3 | x_2)$$

$$= -\sum_{x_1} \sum_{x_2} p(x_1 | x_2) p(x_2) \log p(x_1 | x_2) = -\sum_{x_1} \sum_{x_2} p(x_1, x_2) \log p(x_1 | x_2) := H(1|2) \tag{A.2}$$

$$\therefore H(1|32) \underset{3=f(2)}{\rightarrow} H(1|2)$$